\def\year{2022}\relax
\documentclass[letterpaper]{article} 
\usepackage{aaai22}  
\usepackage{times}  
\usepackage{helvet}  
\usepackage{courier}  
\usepackage[hyphens]{url}  
\usepackage{graphicx} 
\usepackage{natbib}  
\usepackage{caption} 
\DeclareCaptionStyle{ruled}{labelfont=normalfont,labelsep=colon,strut=off} 
\usepackage{algorithm}
\usepackage{algorithmic}

\usepackage{bm}
\usepackage{amssymb}
\usepackage{amsmath}
\usepackage{multirow}
\usepackage{subfigure}
\usepackage[switch]{lineno}

\usepackage{newfloat}
\usepackage{listings}
\floatstyle{ruled}
\newfloat{listing}{tb}{lst}{}
\floatname{listing}{Listing}
\title{Tailor Versatile Multi-modal Learning for Multi-label Emotion Recognition}

\author{Yi Zhang, Mingyuan Chen, Jundong Shen, Chongjun Wang\thanks{Corresponding author}}

\affiliations{
State Key Laboratory for Novel Software Technology, Nanjing University, Nanjing 210023, China \\
\{njuzhangy, mychen, jdshen\}@smail.nju.edu.cn, \{chjwang\}@nju.edu.cn
}

\begin{document}

\maketitle

\begin{abstract}
Multi-modal Multi-label Emotion Recognition (MMER) aims to identify various human emotions from heterogeneous visual, audio and text modalities. 
Previous methods mainly focus on projecting multiple modalities into a common latent space and learning an identical representation for all labels, which neglects the diversity of each modality and fails to capture richer semantic information for each label from different perspectives. 
Besides, associated relationships of modalities and labels have not been fully exploited. 
In this paper, we propose versaTile multi-modAl learning for multI-labeL emOtion Recognition (TAILOR), aiming to refine multi-modal representations and enhance discriminative capacity of each label.
Specifically, we design an adversarial multi-modal refinement module to sufficiently explore the commonality among different modalities and strengthen the diversity of each modality. 
To further exploit label-modal dependence, we devise a BERT-like cross-modal encoder to gradually fuse private and common modality representations in a granularity descent way, as well as a label-guided decoder to adaptively generate a tailored representation for each label with the guidance of label semantics. 
In addition, we conduct experiments on the benchmark MMER dataset CMU-MOSEI in both aligned and unaligned settings, which demonstrate the superiority of TAILOR over the state-of-the-arts. 
Code is available at \url{https://github.com/kniter1/TAILOR}.
\end{abstract}

\section{Introduction}
In real-world applications, videos are often characterized by heterogeneous representations (i.e., visual, audio and text) and annotated with various emotion labels (e.g., \textit{happy}, \textit{surprise}).
Multi-modal Multi-label Emotion Recognition (MMER)~\cite{ju2020transformer,zhang2021multi} refers to identifying various emotions by leveraging visual, audio and text modalities presented in videos.

\begin{figure}[!htbp]
\centering
\includegraphics[width=0.86\linewidth]{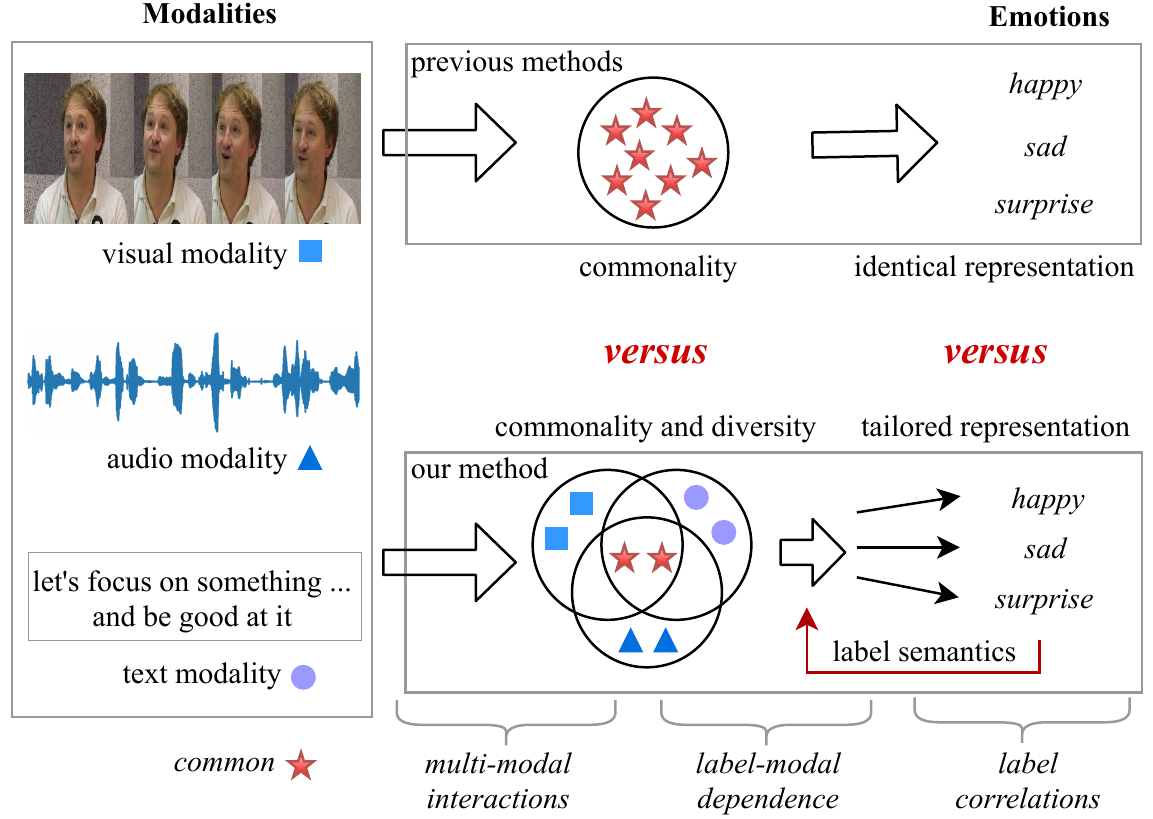} 
\setlength{\belowcaptionskip}{-9pt}
\caption{Previous methods versus our method.}
\label{figVersus}
\end{figure}

Multi-modal learning~\cite{baltrusaitis2019multimodal} processes heterogeneous information collected from multiple sources,  which gives rise to two emergent issues: intra-modal representation and inter-modal fusion.
Intra-modal representation learning mainly exploits consistency and complementarity of multiple modalities to bridge the heterogeneous modality gap.
Previous methods project each modality into a shared latent space to eliminate redundancy.
However, they neglect the fact that different modalities reveal distinctive characteristic of emotions from different perspectives.
Concerning the fusion manner, existing inter-modal fusion methods can be divided into: aggregation-based fusion, alignment-based fusion and the mixture of them~\cite{baltrusaitis2019multimodal}. 
Aggregation-based fusion adopts concatenation~\cite{ngiam2011multimodal}, tensor~\cite{zadeh2017tensor} or attention~\cite{zadeh2018multi} to combine multiple modalities.
Alignment-based fusion centers on latent cross-modal adaptation, which adapts streams from one modality to another~\cite{tsai2019multimodal}.
The key challenge of multi-modal learning lies in 
1) how to integrate commonality while preserving diversity of each individual modality; 
2) how to align different modality distributions interactively for inter-modal fusion.

Multi-label learning~\cite{zhang2014review} deals with rich semantic meanings of complicated objects, where label correlations are considered as the key to effective multi-label learning~\cite{zhu2018multi}.
Many methods exploit label correlations by the similarity between label vectors, and then seamlessly incorporated into an identical representation.
However, they are unable to reflect collaboration relationships among labels.
On the other hand, many researches have been developed to improve the performance by learning label-specific representations in a label-by-label manner, which are generated independently and may lead to suboptimal problem due to the ignorance of label correlations~\cite{zhang2021bilabel}.
The key challenge of multi-label learning is how to effectively encode inherent and discriminative characteristics of each label in both the feature space and the label space.

To address the above challenges, we propose versaTile multi-modAl learning for multI-labeL emOtion Recognition (TAILOR), which sufficiently copes with modality heterogeneity and label heterogeneity.
To bridge the heterogeneity gap, we capture modality interactions, label correlations and label-modal dependence in the following $3$ spaces.

1) \textit{In the modality feature space}, we emphasis less on pre-training.
For intra-modal representation, we devise an adversarial network to explicitly extract commonality and diversity, constrained by common semantics and orthogonality.
For inter-modal fusion, we propose a novel granularity-based fusion with BERT-like transformer encoder.

2) \textit{In the label space}, we adopt self-attention~\cite{vaswani2017attention} to exploit high-order label correlations, which can be further integrated to capture label semantics.

3) \textit{To bridge the gap between modality feature space and label space}, we adapt Transformer decoder to align fused multi-modal representations with label semantics, which aims to learn tailored representation of each label with the guidance of label semantics.

Figure~\ref{figVersus} illustrates the difference between previous methods and our proposed method.
The main contributions can be summarized as follows.
\begin{itemize}
\item A novel framework of versaTile multi-modAl learning for multI-labeL emOtion Recognition (TAILOR) is proposed, which adversarially depicts commonality and diversity among multiple modalities, as well as enhance discriminative capability of label representations.

\item TAILOR adversarially extracts private and common modality representations.
Then a BERT-like transformer encoder is devised to gradually fuse these representations in a granularity descent way, which is incorporated with label semantics to generate tailored label representation.

\item Extensive experiments conducted on benchmark CMU-MOSEI dataset demonstrate the excellent performance of TAILOR in both aligned and unaligned settings.
\end{itemize}


\section{Related Work}
Emotion recognition is broadly studied with \textbf{uni-modal} ~\cite{yang2018sgm,majumder2019dialoguernn,saha2020towards,jiao2020real,huang2021audio}, \textbf{bi-modal} ~\cite{mittal2020emoticon,liu2020federated,zhao2020end} and \textbf{multi-modal}~\cite{mittal2020m3er,sun2020learning,zhang2021multi,lv2021progressive}.
More effective \textbf{multi-modal fusion} translates to better performance.
The most straightforward way is to directly concatenating feature maps from each modality~\cite{ngiam2011multimodal}.
To leverage complementary information across different modalities, tensor fusion~\cite{zadeh2017tensor,liu2018efficient}, memory fusion~\cite{zadeh2018memory}, factorization fusion~\cite{valada2020self} explicitly account for intra-modal and inter-modal dynamics.
The above mentioned methods are aggregation-based fusion and the modality gap heavily affects cross-modal fusion.
To bridge the modality gap, GAN~\cite{goodfellow2014generative} has attracted significant interest in learning joint distributions between bi-modal or multi-modal~\cite{pham2018seq2seq2sentiment,tsai2018learning,pham2019found,mai2020modality}, alignment-based fusion~\cite{baltrusaitis2019multimodal} latently adapts streams from one modality to another via Transformer~\cite{goodfellow2014generative}.
Even though, they tend to fuse into a joint embedding space, which neglects the specificity of each modality.
~\cite{wang2020deep} adapts the fusion of modality-specific streams and fuses only the relevant complementary information.
For example,~\cite{wu2019multi,hazarika2020misa} integrates the common information across modalities, meanwhile preserving the specific patterns of each modality.

In multi-label learning, modeling \textbf{label correlations} has been proven to be an effective strategy~\cite{zhang2014review,zhu2018multi,feng2019collaboration,wang2020collaboration}.
It might be suboptimal to learn a subset of features shared by all the labels.
Another significant strategy is \textbf{label-specific learning}~\cite{zhang2014lift,huang2016learning,zhang2021bilabel}, where each label is determined by some discriminative characteristics, e.g., visual attention~\cite{chen2019learning,chen2019multi} and textual attention~\cite{xiao2019label,ijcai2021CORE}.

Recently, multi-modal multi-label emotion recognition has aroused increasing interest.
For example,~\cite{ju2020transformer,zhang2021multi} models modality-to-label and feature-to-label dependence besides label correlations.

\begin{figure*}[htbp]
\centering
\includegraphics[width=0.8\linewidth]{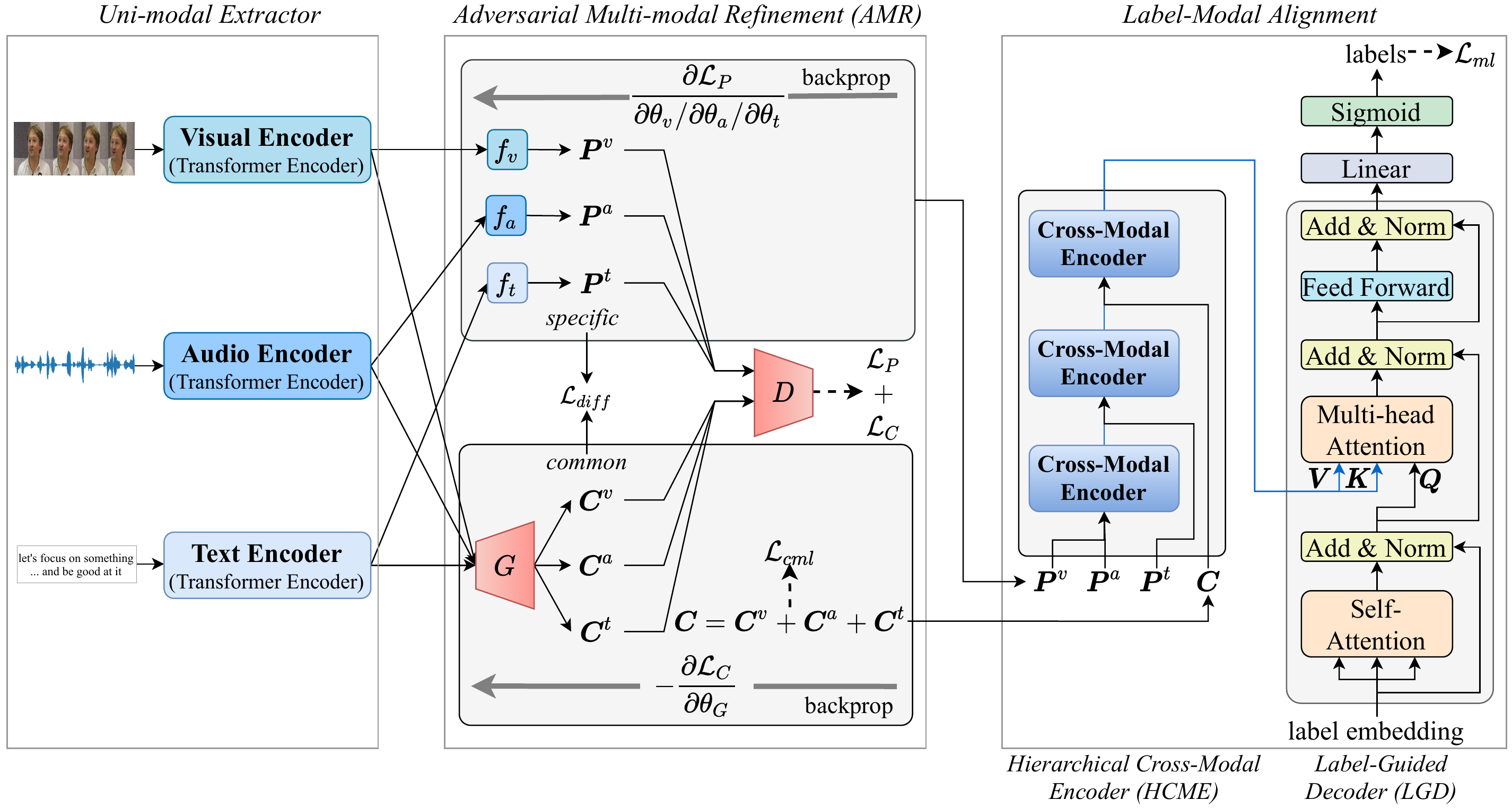} 
\caption{Overall structure of TAILOR. }
\label{figFlowchart}
\end{figure*}
\section{Methodology}
In this section, we first give the formulation of Multi-modal Multi-label Emotion Recognition (MMER).
We use lowercase for scalars (e.g., $a$), uppercase for vectors (e.g., $A$) and bold for matrices (e.g., $\bm a$, $\bm A$).
Let $\mathcal X^{v} = \mathbb R^{d_{v} \times \tau_{v}}$, $\mathcal X^{a} = \mathbb R^{d_{a} \times \tau_{a}}$, $\mathcal X^{t} = \mathbb R^{d_{t} \times \tau_{t}}$ be the visual ($v$), audio ($a$), text ($t$) feature space respectively, and $\mathcal Y = \{y_{1}, y_{2}, \cdots, y_{l}\}$ denote the label space with $l$ labels, where $d_{\{v, a, t\}}$ represents modality dimension and $\tau_{\{v, a, t\}}$ represents sequence length.
Given a training dataset $\mathcal D = \{(\bm X_{i}^{\{v, a, t\}}, Y_{i})\}_{i=1}^{n}$ with $n$ data samples, the goal of MMER is to learn a function $\mathcal F: \mathcal X^{v} \times \mathcal X^{a} \times \mathcal X^{t} \rightarrow 2^{\mathcal Y}$, which can assign a set of possible emotion labels for the unseen video.
For the $i$-th video, $\bm X_{i}^{\{v, a, t\}} \in \mathcal X^{\{v, a, t\}}$ is the modality features and $Y_{i} \subseteq \mathcal Y$ is the set of relevant labels.
Fig.~\ref{figFlowchart} shows the main framework of TAILOR, which comprises the following modules: \textit{Uni-modal Extractor}, \textit{Adversarial Multi-modal Refinement} and \textit{Label-Modal Alignment}.
\subsection{Uni-modal Extractor}
The pre-extracted features for each modality in CMU-MOSEI~\cite{zadeh2018multimodal} dataset are represented by asynchronous coordinated sequences.
To exploit long-term contextual information, we use $n_{v}$-layer, $n_{a}$-layer, $n_{t}$-layer Transformer encoder~\cite{vaswani2017attention} to enrich the visual features, audio features and text features with sequence level context separately.
The transformer encoder consists of two sub-layers: multi-head self-attention layer and position-wise feed-forward layer, where residual connections~\cite{he2016deep} are adopted, followed by layer normalization.
As a result, we obtain new visual, audio and text embeddings, denoted as $\bm V \in \mathbb R^{d \times \tau}, \bm A \in \mathbb R^{d \times \tau}, \bm T \in \mathbb R^{d\times \tau}$, where $d$ is the embedding dimension and $\tau$ is the sequence length.

\subsection{Adversarial Multi-modal Refinement}
It is well known that the greater the difference between inter-modal representations, the better the complementarity of inter-modal fusion~\cite{yu2020ch}.
Though the uni-modal extractors capture long-term temporal context, they are unable to deal with feature redundancy due to modality gap.
Inspired by adversarial networks~\cite{goodfellow2014generative}, we design an adversarial multi-modal refinement module for the subsequent fusion.
It inherently decomposes multiple modalities to two disjoint parts: common and private representations so as to extract consistency and specificity of heterogeneous modalities collaboratively and individually.

To maintain consistency, we design a generator $G(\cdot; \theta_{G})$ with parameters $\theta_{G}$, to project different modalities into a common latent subspace with distributional alignment.
Apart from commonality, each modality contains specific information, which can complement with other modalities.
We adopt fully connected deep neural networks $f_{v}(\cdot; \theta_{v})$, $f_{a}(\cdot; \theta_{a})$ and $f_{t}(\cdot; \theta_{t})$ with parameters $\{\theta_{v}$, $\theta_{a}$, $\theta_{t}\}$ to project uni-modal embedding $\{\bm V, \bm A, \bm T\}$ respectively.
The common and private representations can be formulated as,
\begin{equation}
\begin{split}
\bm C^{v} &= G(\bm V; \theta_{G}), \ \bm C^{a} = G(\bm A; \theta_{G}), \ \bm C^{t} = G(\bm T; \theta_{G}) \\
\bm P^{v} &= f_{v}(\bm V, \theta_{v}), \ \bm P^{a} = f_{a}(\bm A; \theta_{a}), \ \bm P^{t} = f_{t}(\bm T, \theta_{t})
\end{split}
\end{equation}
where $\bm C^{\{v, a, t\}},\bm P^{\{v, a, t\}} \in \mathbb R^{d\times \tau}$.

\noindent\textbf{Adversarial Training}
In order to guarantee the purity of common and specific representations, we design a modality discriminator $D(,;\theta_{D})$ which maps the input $\bm I \in \mathbb R^{d\times \tau}$ into a probability distribution and estimates which modality the representation comes from, where $d$ is the modality dimension and $\tau$ is the sequence length.
\begin{equation}
\begin{split}
& D(\bm I; \theta_{D}) = \text{softmax}(\bm I^{T} \bm W + \bm b) \\
\end{split}
\end{equation}
where $\bm W \in \mathbb R^{d\times 3}$ is the weight matrices, and $\bm b \in \mathbb R^{\tau\times 3}$ is the bias matrices.
The ground truth modality label of $\bm I$ is denoted as $\bm O \in \{\bm O^{v}, \bm O^{a}, \bm O^{t}\}$,
\begin{equation}
\bm O^{v}=
\left[
\begin{aligned}
& 1, 0, 0 \\
& \ \ \cdots \\
& 1, 0, 0\\
\end{aligned}
\right],
\bm O^{a}=
\left[
\begin{aligned}
& 0, 1, 0 \\
& \ \ \cdots \\
& 0, 1, 0\\
\end{aligned}
\right],
\bm O^{t}=
\left[
\begin{aligned}
& 0, 0, 1 \\
& \ \ \cdots \\
& 0, 0, 1\\
\end{aligned}
\right],
\end{equation}
where $\bm O^{v}, \bm O^{a}, \bm O^{t}\in \mathbb R^{\tau \times 3}$.

{Common representations} $\bm C^{\{v, a, t\}}$ are encoded in a shared latent subspace, which tends to be in the same distribution.
Therefore, the generator $G(,;\theta_{G})$ are encouraged to confuse discriminator $D(,; \theta_{D})$ thus not to distinguish the source modality of $\bm C^{\{v, a, t\}}$.
We reconstruct a training dataset $\mathcal D_{C} = \{(\bm C_{i}^{v}, \bm O^{v}), (\bm C_{i}^{a}, \bm O^{a}), (\bm C_{i}^{t}, \bm O^{t})\}_{i=1}^{n}$ for common modality classification.
The common adversarial loss is,
\begin{equation}
\mathcal L_{C} = -\frac{1}{n}\sum_{m\in\{v, a, t\}}\sum_{i=1}^{n}(\bm O^{m}\text{log}(D(\bm C_{i}^{m}; \theta_{D})))
\end{equation}
where $\mathcal L_{C}$ is trained with gradient reversal layer~\cite{ganin2015unsupervised} that leaves the input unchanged during forward propagation and multiply the gradient by $-1$ during the backpropagation.

{Private representations} $\bm P^{\{v, a, t\}}$ are encoded in diverse latent subspaces, which tends to be in different distributions.
Therefore, the discriminator $D(\cdot; \theta_{D})$ are encouraged to distinguish the source of modality.
We reconstruct a training dataset $\mathcal D_{P} = \{(\bm P_{i}^{v}, \bm O^{v}), (\bm P_{i}^{a}, \bm O^{a}), (\bm P_{i}^{t}, \bm O^{t})\}_{i=1}^{n}$ for private modality classification.
The private adversarial loss is,
\begin{equation}
\mathcal L_{P} = -\frac{1}{n}\sum_{m\in\{v, a, t\}}\sum_{i=1}^{n}(\bm O^{m}\text{log}(D(\bm P_{i}^{m}; \theta_{D})))
\end{equation}

\noindent\textbf{Orthogonal Constraint}
To encode different aspects of multi-modal data, we penalize redundancy in $\bm C^{\{v, a, t\}}$ and $\bm P^{\{v, a, t\}}$ with orthogonal loss as follows, 
\begin{equation}
\mathcal L_{diff} = - \sum_{m\in \{v, a, t\}}\sum_{i=1}^{n}||(\bm C_{i}^{m})^{T}\bm P_{i}^{m}||_{F}^{2}
\end{equation}
where $||\cdot||_{F}^{2}$ is the squared Frobenius norm.

\noindent\textbf{Common Semantics}
Although the common generator $G(,; \theta_{G})$ and private extractors $f_{v}(,; \theta_{v})$, $f_{a}(,; \theta_{a})$ $f_{t}(,; \theta_{t})$ are encouraged to encode different aspects of multi-modal information, they should exhibit the same semantics.
We are motivated to design common semantic loss for multi-label classification with common representations $\bm C^{{\{v, a, t\}}}$, 
\begin{equation}
\mathcal L_{cml} = - \sum_{m\in \{v, a, t\}}\sum_{i=1}^{n}\sum_{j=1}^{l}y_{i}^{j}\text{log}\hat y_{i}^{j, m} + (1-y_{i}^{j})\text{log}(1-\hat y_{i}^{j, m}))
\end{equation}
where $\hat y_{i}^{j, m}$ is predicted with $\bm C^{m}$ and $y_{i}^{j}$ is the ground-truth.
$y_{i}^{j}=1$ if the $j$-th label is relevant, $0$ otherwise.

\subsection{Label-modal Alignment}
After projecting into private and common representations respectively and collectively, we need to fuse them into a joint representation for multi-label classification.

\noindent\textbf{Hierarchical Cross-Modal Encoder}
The refined common and private modality representations contain consistent and complementary information, while few or no information with regard to modality interactions.
Simply concatenating them together ignores modality interactions, which might introduce redundant information and lead to suboptimal problem~\cite{zhang2018latent}.
We propose a novel BERT-like~\cite{kenton2019bert} Cross-Modal Encoder to exploit modality interactions.

\begin{figure}[htbp]
\centering
\includegraphics[width=0.9\linewidth]{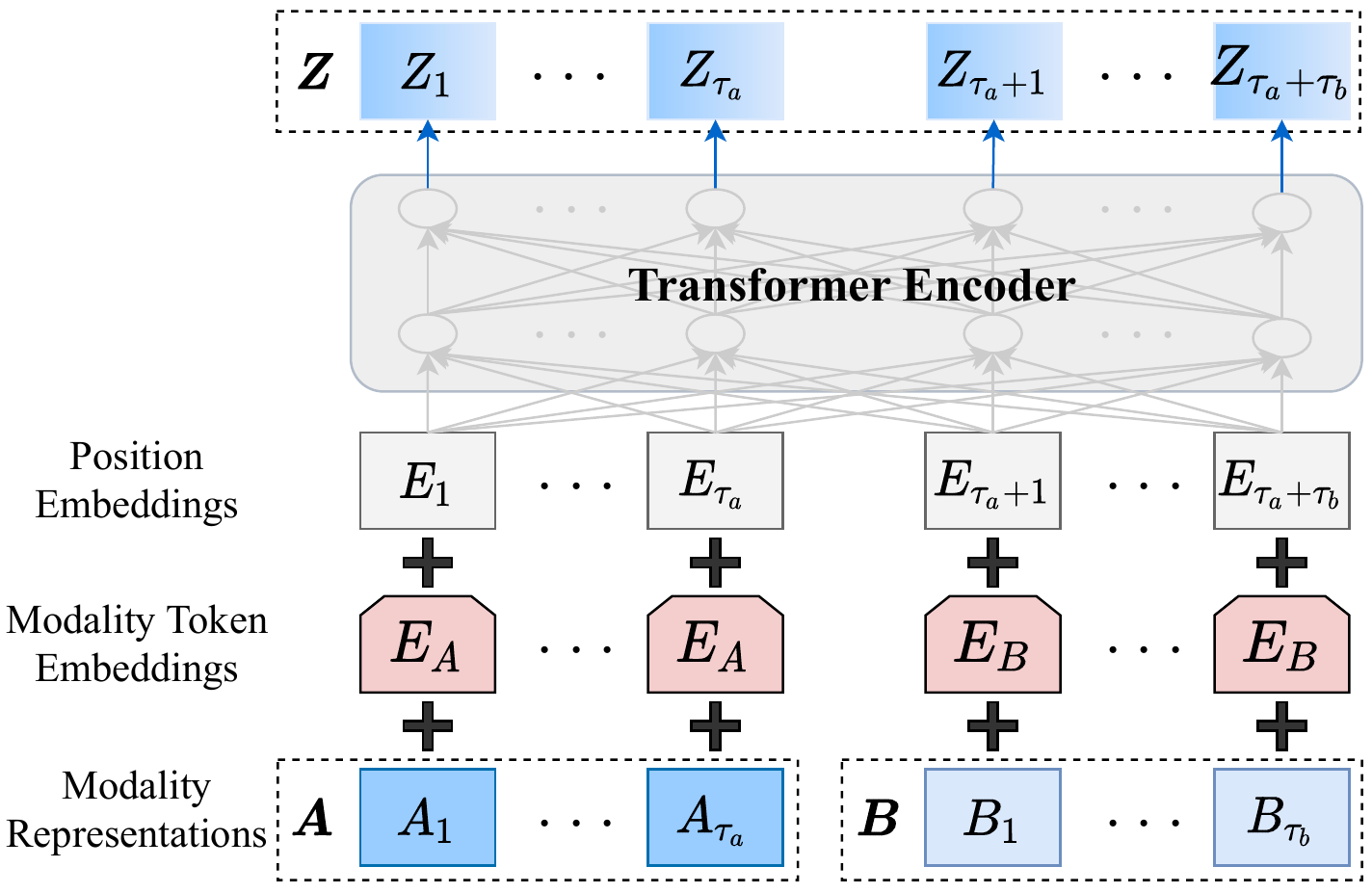} 
\caption{Structure of Cross-Modal Encoder (CME) between two modalities $\bm A$ and $\bm B$. 
The two input modality representations, modality token embeddings and position embeddings are summed and fed into the Transformer encoder.}
\label{figCrossEncoder}
\end{figure}

Given two modalities $a$ and $b$ with representations $\bm A \in \mathbb R^{d\times \tau_{a}}$ and $\bm B \in \mathbb R^{d\times \tau_{b}}$, where $d$ is modality dimension and $\tau_{\{a, b\}}$ is sequence length.
On the one hand, to preserve the temporal information of the two modalities, we augment them with positional embeddings $\bm E \in \mathbb R^{d\times (\tau_{a}+\tau_{b})}$.
On the other hand, the feature distribution of various modalities are different due to heterogeneity, which poses a great challenge to multi-modal fusion.
To bridge the large margin of the statistical properties between two modalities, we capture statistical regularities by adding two modality token embeddings $E_{A}\in \mathbb R^{1\times \tau_{a}}$ and $E_{B}\in \mathbb R^{1\times \tau_{b}}$ to modality $a$ and $b$ respectively.
As illustrated in Fig.~\ref{figCrossEncoder}, the sum of modality representations, position embeddings and modality token embeddings is feed into $n_{c}$-layer Transformer Encoder, which outputs the joint representation $\bm Z \in \mathbb R^{d\times (\tau_{a}+\tau_{b})}$ of modality $a$ and $b$.
Cross-Modal Encoder can be written as $\bm Z = \text{CME}(\bm A, \bm B)$.

Besides, visual and audio modalities are more fine-grained than text modality in terms of granularity~\cite{alayrac2020self}, which is rarely considered in existing fusion methods.
To remedy the deficiency, we devise Hierarchical Cross-Modal Encoder (HCME) to exploit interactions across modalities with different level of granularities.
Private representations $\bm P^{\{v, a, t\}}$ and common representations $\bm C^{\{v, a, t\}}$ are fused in a hierarchical structure and gradually complement with each other in a granularity descent way.
\begin{equation}
\begin{split}
& \text{Fine-grained} \ \ \ \  \ \ \bm Z^{va} = \text{CME}(\bm P^{v}, \bm P^{a}) \\
& \text{Coarse-grained} \ \ \bm Z^{vat} = \text{CME}(\bm Z^{va}, \bm P^{t}) \\
& \text{Mixed-grained} \ \ \ \bm M = \text{CME}(\bm Z^{vat}, \bm C) \\
\end{split}
\end{equation}
where $\bm C=\bm C^{v}+\bm C^{a}+\bm C^{t} \in \mathbb R^{d\times \tau}$, $\bm M \in \mathbb R^{d\times 4\tau}$.
HCME models $3$ pairs of modalities with the fusion order $\psi=[v, a, t, c]$.
Each pair of modalities interacts and correlates valuable information step by step.

\noindent\textbf{Label-Guided Decoder}
Label correlations plays an important role in effective multi-label classification.
For $l$ possible labels in original label space $\bm Y = [Y_{1}, Y_{2}, \cdots, Y_{n}] \in \mathbb R^{l \times n}$, we use label indices to produce the label embedding $\bm L=[L_{1}; L_{2}; \cdots; L_{l}] \in \mathbb R^{l \times d}$ where $l$ is the number of labels and $d$ is label dimension which is the same as modality dimension.
$\widetilde k = \{1, 2, \cdots, l\}\backslash k$ denotes all labels except the $k$-th label.
$\bm L_{k} \in \mathbb R^{1\times d}$ is the label embedding for the $k$-th label, while $\bm L_{\widetilde k} = [L_{1}; \cdots; L_{k-1}; L_{k+1}; \cdots; L_{l}] \in \mathbb R^{(l-1)\times d}$ is the label embedding for $\widetilde k$.
To exploit label correlations collaboratively, we adopt self-attention mechanism with $h_{l}$ heads.
For the $i$-th head \footnote{we use bold lowercase to denote parameters in each head},
\begin{equation}
\begin{split}
& \bm q = \bm L \bm W_{i}^{q} =
\left[
\begin{aligned}
\bm L_{k} \\
\bm L_{\widetilde k}
\end{aligned}
\right]
\bm W_{i}^{q}
=
\left[
\begin{aligned}
\bm q_{k} \\
\bm q_{\widetilde k}
\end{aligned}
\right]
, \bm W_{i}^{q} \in \mathbb R^{d \times d/h_{l}}
\end{split}
\end{equation}
Similarly,
$
\bm k =
\left[
\begin{aligned}
\bm k_{k} \\
\bm k_{\widetilde k}
\end{aligned}
\right]
,
\bm v = 
\left[
\begin{aligned}
\bm v_{k} \\
\bm v_{\widetilde k}
\end{aligned}
\right]
$, $\bm W_{i}^{k}, \bm W_{i}^{v} \in \mathbb R^{d \times d/h_{l}}$.

And then label correlation matrix $\bm r$ can be calculated as,
\begin{equation}
\begin{split}
\bm r = \bm q\bm k^{T}
=
\left[
\begin{aligned}
\bm q_{k}\bm k_{k}^{T}\quad \bm q_{k}\bm k_{\widetilde k}^{T} \\
\bm q_{\widetilde k}\bm k_{k}^{T}\quad \bm q_{\widetilde k}\bm k_{\widetilde k}^{T}
\end{aligned}
\right]
=
\left[
\begin{aligned}
\bm r_{kk} \quad \bm r_{k\widetilde k} \\
\bm r_{\widetilde kk} \quad \bm r_{\widetilde k\widetilde k}
\end{aligned}
\right]
\in \mathbb R^{l\times l}
\end{split}
\end{equation}
where $\bm r_{{kk}}$ and $\bm r_{\widetilde k\widetilde k}$ represent the label-specific relation, $\bm r_{{k\widetilde k}}$ and $\bm r_{\widetilde kk}$ represent the interactive relation of the $k$-th label with respect to other $l-1$ labels.
$\bm r_{{k\widetilde k}}$ denotes the influence of other $l-1$ labels to the $k$-th label, while $\bm r_{{\widetilde kk}}$ denotes the influence of the $k$-th label to other $l-1$ labels.
The label semantic embedding $\bm S$ of the $i$-th head is,
\begin{equation}
\begin{split}
\bm S_{i} =
\left[
\begin{aligned}
\bm s_{k} \\
\bm s_{\widetilde k}
\end{aligned}
\right]
& =\text{softmax}\big(\frac {1}{\sqrt{d/h_{l}}}
\left[
\begin{aligned}
\bm r_{kk} \quad \bm r_{k\widetilde k} \\
\bm r_{\widetilde kk} \quad \bm r_{\widetilde k\widetilde k}
\end{aligned}
\right]\big)
\left[
\begin{aligned}
\bm v_{k}  \\
\bm v_{\widetilde k}
\end{aligned}
\right]
\\
&=
\left[
\begin{aligned}
\sigma(\bm r_{kk})\bm v_{k} + \sigma(\bm r_{k\widetilde k})\bm v_{\widetilde k} \\
\sigma(\bm r_{\widetilde kk})\bm v_{k} + \sigma(\bm r_{\widetilde k\widetilde k})\bm v_{\widetilde k}
\end{aligned}
\right]
\in \mathbb R^{l\times d/h_{l}}
\end{split}
\label{r}
\end{equation}
where $\sigma(\bm r) = \text{softmax}(\frac {\bm r}{\sqrt{d/h_{l}}})$ is a row-wise, scaled softmax.
For the $k$-th label, the label-specific semantic embedding is $\bm s_{k} = \sigma(\bm r_{kk})\bm v_{k} + \sigma(\bm r_{k\widetilde k})\bm v_{\widetilde k}$, which involves the collaboration of its own semantic implication and the semantic implication receiving from other labels.
In addition, we add a residual connection followed by layer normalization (LN), to the final label-specific semantic embeddings,
\begin{equation} 
\begin{split}
& \bm L = \text{LN}(\bm L + \bm S) \\
& \bm S = \text{Concat}(\bm S_{1}, \bm S_{2}, \cdots, \bm S_{h_{l}}) \bm W^{L}
\end{split}
\end{equation}
where $\bm W^{L}\in \mathbb R^{d\times d}$, $\bm S \in \mathbb R^{l\times d}$.

Label semantics determine inherent dependence between labels and modalities.
Therefore, the obtained label-specific semantic embeddings $\bm L \in \mathbb R^{l\times d}$ can be further considered as a teacher to guide the learning of tailored representation for each label.
Inspired by transformer decoder~\cite{vaswani2017attention}, we design a label-guided decoder to select discriminative information from joint multi-modal representations $\bm M \in \mathbb R^{d\times 4\tau}$ with the guidance of label semantics.
The latent dependence from modality space to label space is captured by multi-head attention with $h_{m}$ heads,
\begin{equation}
\begin{split}
\bm {Dep}_{M\rightarrow L} = \text{Concat}(\text{dep}_{1}, \cdots, \text{dep}_{{h_{m}}})\bm W^{M} \\
\ \text{dep}_{i} = \text{softmax}(\frac {\bm L\bm W_{i}^{Q}(\bm M^{T}\bm W_{i}^{K})^{T}}{\sqrt{d /h_{m}}})\bm M^{T}\bm W_{i}^{V}
\end{split}
\end{equation}
where $\bm W_{i}^{Q}, \bm W_{i}^{K}, \bm W_{i}^{V} \in \mathbb R^{d \times d/h_{m}}$, $\text{dep}_{i} \in \mathbb R^{l\times d/h_{m}}$, $\bm W^{M} \in \mathbb R^{d \times d}$, $\bm {Dep}_{M\rightarrow L} \in \mathbb R^{l\times d}$.

Then the tailored representations $\bm H =[H_{1}; \cdots; H_{l}]\in \mathbb R^{l\times d}$ is generated by a feed-forward network (FFN) and two layer normalization (LN) with residual connection.
\begin{equation}
\begin{split}
\hat{\bm L} = \text{LN}(\bm L + \bm {Dep}_{M\rightarrow L})\\
\bm H = \text{LN}(\hat{\bm L} + \text{FFN}(\hat{\bm L})
\end{split}
\end{equation}

\noindent\textbf{Multi-label Classification} For the $k$-th label, its tailored representation $H_{k}$ is fed into a linear function followed by an output sigmoid for the final label classification,
\begin{equation}
\label{EqY}
\mathcal F_{k} = \text{sigmoid}(H_{k}W_{k}+b_{k})
\end{equation}
where $W_{k} \in \mathbb R^{d}$ is weight vector and $b_{k} \in \mathbb R$ is the bias.\\
The final multi-label classification loss can be computed with binary cross-entropy loss,
\begin{equation}
\mathcal L_{ml} = -\sum_{i=1}^{n}\sum_{j=1}^{l}y_{i}^{j}\text{log}\hat y_{i}^{j} + (1-y_{i}^{j})\text{log}(1-\hat y_{i}^{j})
\end{equation}
where $\hat y_{i}^{j}$ is predicted by Eq.~\ref{EqY} and $y_{i}^{j}$ is the ground-truth.
$y_{i}^{j}=1$ if the $j$-th label is relevant, $0$ otherwise.

Above all, combining the final multi-label classification loss $\mathcal L_{ml}$, common adversarial loss $\mathcal L_{C}$, private adversarial loss $\mathcal L_{P}$,  common semantic loss $\mathcal L_{cml}$ and orthogonal loss $\mathcal L_{diff}$ together, 
the final objective function is computed as,
\begin{equation}
\mathcal L_{All} = \mathcal L_{ml} + \alpha ( \mathcal L_{C} + \mathcal L_{P}) + \beta \mathcal L_{diff} + \gamma \mathcal L_{cml}
\end{equation}
where $\alpha$, $\beta$ and $\gamma$ are the trade-off parameters.

\section{Experiment}
In this section, we give empirically evaluations and analysis of our proposed TAILOR method.
\subsection{Experimental Setup}
\noindent\textbf{Dataset}
We conduct experiments on benchmark multi-modal multi-label dataset CMU-MOSEI~\footnote{https://github.com/A2Zadeh/CMU-MultimodalSDK}~\cite{zadeh2018multimodal}, which contains $22,856$ video segments from $1,000$ distinct speakers.
Each video inherently contains $3$ modalities: visual, audio and text, while annotated with $6$ discrete emotions: \textit{\{angry, disgust, fear, happy, sad, surprise\}}.
We pre-extract $35$-dimensional visual features from video frames by FACET~\cite{baltruvsaitis2016openface}, $74$-dimensional audio features from acoustic signals by COVAREP~\cite{degottex2014covarep} and $300$-dimensional text features from video transcripts by Glove~\cite{pennington2014glove}.
Table~\ref{TableData} summarizes details of CMU-MOSEI in both word-aligned and unaligned settings.

\begin{table}[htbp]
\caption{Statistics of CMU-MOSEI, where $d_{\{v, a, t\}}$ is modality dimension and $\tau_{\{v, a, t\}}$ is sequence length.}
\centering
\resizebox{.99\linewidth}{!}{
\begin{tabular}{c|cccc|ccc|ccccc}
\hline

\hline
& \#instance & \#train & \#valid & \#test & $d_{v}$ & $d_{a}$ & $d_{t}$ & $\tau_{v}$ & $\tau_{a}$ & $\tau_{t}$ \\
\hline

\hline
aligned 
& \multirow{2}{*}{$22856$} & \multirow{2}{*}{$16326$} & \multirow{2}{*}{$1871$} & \multirow{2}{*}{$4659$}
& \multirow{2}{*}{$35$} & \multirow{2}{*}{$74$} & \multirow{2}{*}{$300$}
& $60$ & $60$ & $60$ \\
unaligned 
& & & & & & & & $500$ & $500$ & $50$\\
\hline

\hline
\end{tabular}
}
\label{TableData}
\end{table}

\begin{table*}[htb]
\caption{Predictive performance of TAILOR on multi-modal multi-label CMU-MOSEI dataset with aligned and unaligned multi-modal sequences compared with state-of-the-arts. The best performance for each criterion is bolded.}
\centering
\resizebox{.8\textwidth}{!}{
\begin{tabular}{c|cccc|cccc}
\hline

\hline
\multirow{2}{*}{Approaches} & \multicolumn{4}{c|}{Aligned} & \multicolumn{4}{c}{Unaligned}\\
\cline{2-9}
& Acc & P & R & Micro-F1 & Acc & P & R & Micro-F1 \\
\hline

\hline
BR~\cite{boutell2004learning}
& 0.222 & 0.309 & 0.515 & 0.386 
& 0.233 & 0.321 & 0.545 & 0.404 \\
LP~\cite{tsoumakas2007multi} 
& 0.159 & 0.231 & 0.377 & 0.286
& 0.185 & 0.252 & 0.427 & 0.317 \\
CC~\cite{read2011classifier}
& 0.225 & 0.306 & 0.523 & 0.386
& 0.235 & 0.320 & 0.550 & 0.404 \\
\hline

\hline
SGM~\cite{yang2018sgm}
& 0.455 & 0.595 & 0.467 & 0.523 
& 0.449 & 0.584 & 0.476 & 0.524 \\
LSAN~\cite{xiao2019label}
& 0.393 & 0.550 & 0.459 & 0.501
& 0.403 & 0.582 & 0.460 & 0.514 \\
ML-GCN~\cite{chen2019multi}
& 0.411 & 0.546 & 0.476 & 0.509
& 0.437 & 0.573 & 0.482 & 0.524 \\
\hline

\hline
DFG~\cite{zadeh2018multimodal}
& 0.396 & 0.595 & 0.457 & 0.517
& 0.386 & 0.534 & 0.456 & 0.494 \\
RAVEN~\cite{wang2019words}
& 0.416 & 0.588 & 0.461 & 0.517
& 0.403 & 0.633 & 0.429 & 0.511 \\
MulT~\cite{tsai2019multimodal}
& 0.445 & 0.619 & 0.465 & 0.531 
& 0.423 & 0.636 & 0.445 & 0.523 \\
SIMM~\cite{wu2019multi} 
& 0.432 & 0.561 & 0.495 & 0.525
& 0.418 & 0.482 & 0.486 & 0.484 \\
MISA~\cite{hazarika2020misa}
& 0.430 & 0.453 & \textbf{0.582} & 0.509
& 0.398 & 0.371 & \textbf{0.571} & 0.450 \\
HHMPN~\cite{zhang2021multi} 
& 0.459 & 0.602 & 0.496 & 0.556
& 0.434 & 0.591 & 0.476 & 0.528 \\
\hline

\hline
TAILOR 
& \textbf{0.488} & \textbf{0.641} & 0.512 & \textbf{0.569}
& \textbf{0.460} & \textbf{0.639} & 0.452 & \textbf{0.529} \\
\hline

\hline
\end{tabular}
}
\label{TableExp}
\end{table*}

\noindent\textbf{Evaluation Metrics}
We adopt $4$ mostly used multi-label classification evaluation metrics~\cite{zhang2014review}: Accuracy (Acc), Micro-F1, Precision (P) and Recall (R).
Larger value indicates better performance.

\noindent\textbf{Compared Approaches}
On the one hand, we conduct experiments with Multi-Label Classification (MLC) methods.
For classic methods \textbf{BR}~\cite{boutell2004learning}, \textbf{LP}~\cite{tsoumakas2007multi} and \textbf{CC}~\cite{read2011classifier}, we concatenate all modalities as a new input.
For text/image based methods, we only use text/image modality.
\textbf{SGM}~\cite{yang2018sgm} views MLC as a sequence generation problem to take label correlations into account.
\textbf{LSAN}~\cite{xiao2019label} considers document content and label texts simultaneously.
\textbf{ML-GCN}~\cite{chen2019multi} captures label correlations for multi-label image recognition and employs GCN to map label representations.

On the other hand, we compare with multi-modal multi-label methods.
\textbf{DFG}~\cite{zadeh2018multimodal} studies the nature of cross-modal dynamics in multimodal language.
\textbf{RAVEN}~\cite{wang2019words} captures dynamic nature of nonverbal intents by shifting word representations based on the accompanying nonverbal behaviors.
\textbf{MulT}~\cite{tsai2019multimodal} fuses multi-modal information by directly attending to low-level features in other modalities.
\textbf{SIMM}~\cite{wu2019multi} leverages shared subspace exploitation and view-specific information extraction with adversarial learning.
\textbf{MISA}~\cite{hazarika2020misa} learns modality-invariant and -specific representations as a pre-cursor to multi-modal fusion. 
\textbf{HHMPN}~\cite{zhang2021multi} simultaneously models feature-to-label, label-to-label and modality-to-label dependencies via graph message passing.

\noindent\textbf{Implementation Details}
We set hyper-parameters $\alpha=0.01$, $\beta=5e{-6}$ and $\gamma=0.5$.
The batch size is 64.
For layer number in Transformer Encoder, we set $n_{v}=n_{a}=4$, $n_{t}=6$ in uni-modal encoders,
$n_{c}=3$ in cross-modal encoders.
The size of hidden layers in encoders and decoder is $d=256$, the head number $h_{l}=h_{m}=8$.
All parameters in TAILOR are optimized by Adam~\cite{kingma2015adam} with an initial learning rate of $1e{-5}$ for aligned setting, $1e{-4}$ for unaligned setting and employ a liner decay learning rate schedule with a warm-up strategy.
All experiments are running with one GTX 1080Ti GPU.

\subsection{Experimental Results and Analysis}
\noindent\textbf{Experimental Results}
Except MulT, we include CTC ~\cite{graves2006connectionist} to be suitable for the unaligned setting.
Based on the comparison results in Table~\ref{TableExp}, we have the following observations.
1) Our proposed TAILOR significantly surpasses the state-of-the art methods on all evaluation metrics except recall (R), which is relatively less important than accuracy (Acc) and Micro-F1 for performance evaluation.
2) \textbf{CC} performs best among $3$ classic multi-label methods, which indicates the effectiveness of exploiting label correlations.
3) Text based multi-label methods \textbf{SGM}, \textbf{LSAN} and image based multi-label methods \textbf{ML-GCN} performs better than \textbf{CC}, which further conforms that label correlations conduce to capture more meaningful features.
4) \textbf{MulT} performs better than almost all multi-label methods that with concatenated modalities or only with text/image modality, which shows the necessity of exploiting modality complementarity.
5) Multi-modal multi-label methods such as \textbf{HHMPN} performs even better than aforementioned methods, which validates the effectiveness of exploiting modality information and label information simultaneously.

\noindent\textbf{Ablation Study}
To get a better understanding of TAILOR, we investigate different components in $3$ main modules: AMR, HCME, LGE.
\begin{table}[htb]
\caption{Ablation experiments of TAILOR on the aligned CMU-MOSEI dataset. ``w/o'' denotes removing the component, ``w/ '' denotes adding the component, $\psi$ is fusion order. MTE: Modality Token Embedding, LE: Label Embedding, LC: Label Correlations. ``w/ identical'' denotes prediction with fused modalities via a dense layer.}
\centering
\resizebox{.91\linewidth}{!}{
\begin{tabular}{l|cccc}
\hline

\hline
{Approaches} & Acc & P & R & Micro-F1 \\
\hline
(1) w/o AMR
& 0.446 & 0.634 & 0.474 & 0.543 \\
(2) w/ $\mathcal L_{adv}$
& 0.432 & 0.722 & 0.419 & 0.530 \\
(3) w/ $\mathcal L_{adv}$, $\mathcal L_{diff}$
& 0.462 & 0.581 & 0.520 & 0.549 \\
\hline
(4) w/ $\bm C^{\{v, a, t\}}$
& 0.458 & 0.638 & 0.481 & 0.549 \\
(5) w/ $\bm P^{\{v, a, t\}}$
& 0.449 & 0.605 & 0.496 & 0.545 \\
(6) $\psi = [v, t, a, c]$
& 0.465 & 0.629 & 0.496 & 0.554 \\
(7) $\psi = [a, t, v, c]$
& 0.470 & 0.584 & 0.524 & 0.552 \\
(8) w/o MTE
& 0.478 & 0.601 & 0.528 & 0.562 \\
\hline
(9) w/ identical
& 0.462 & 0.575 & 0.528 & 0.551 \\
(10) w/ LE 
& 0.465 & 0.558 & \textbf{0.556} & 0.557 \\
(11) w/ LE,  LC
& 0.473 & 0.594 & 0.538 & 0.564 \\
\hline
(12) TAILOR 
& \textbf{0.488} & \textbf{0.641} & 0.512 & \textbf{0.569} \\

\hline

\hline
\end{tabular}
}
\label{TableAblation}
\end{table}
Ablation results are shown in Table~\ref{TableAblation} and several observations are obtained as follows.
\begin{itemize}
\item (1) is worst, which validates significance of adversarial multi-modal learning.
As integrating $\mathcal L_{adv}$, $\mathcal L_{diff}$, $\mathcal L_{cml}$ to AMR optimization, (2), (3), (12) gradually improves.
\item (4), (5) are worse than (12), which reveals that joint consideration of the commonality and diversity of multi-modal data leads to better performance.
\item Changing the fusion order of HCME leads to poor performance, while (6), (7) is better than (1)-(5).
It validates the rationality and optimality of HCME.
\item (8) is worse than (12), where modality token embedding can really help bridge low-level modality gap.
\item Label-specific results (10)-(12) gradually get better, which are all better than (9). 
Effective label-specific learning with label correlations and label-modal dependence can enhance discriminative power of each label.
\end{itemize}

\noindent\textbf{Effect of Adversarial Learning}
In the Adversarial Muti-modal Refinement (AMR) module, we jointly optimize common adversarial loss, private adversarial loss, common semantic loss, orthogonal loss and overall loss.
\begin{figure}[htb]
\centering
\includegraphics[width=0.9\linewidth]{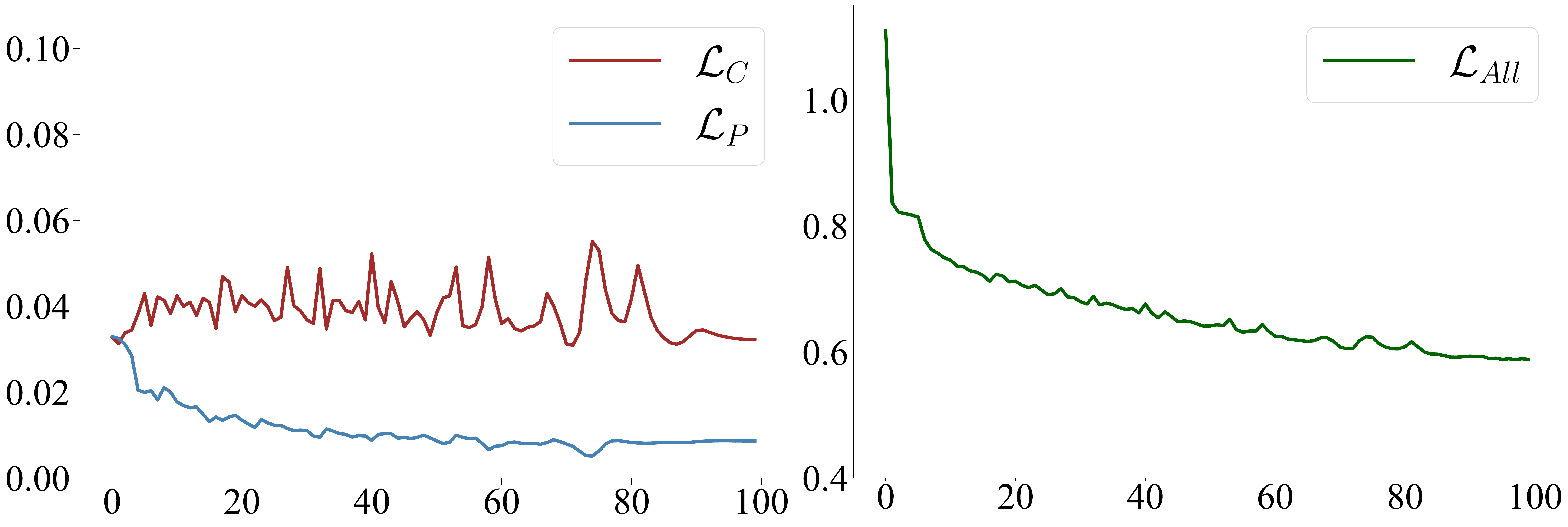} 
\caption{Common adversarial loss $\mathcal L_{C}$, private adversarial loss $\mathcal L_{P}$ and overall loss $\mathcal L_{All}$ w.r.t. the number of epochs.}
\label{figAdvLoss}
\end{figure}
As shown in Figure~\ref{figAdvLoss}, private adversarial loss $\mathcal L_{P}$ and overall loss $\mathcal L_{All}$ decreases almost monotonously and converges smoothly, while common adversarial loss $\mathcal L_{C}$ first vibrates and later tends to be stable.
In the end, it reaches a point where neither common nor private adversarial learning can improve.

\begin{figure}[htb]
\centering
\includegraphics[width=1\linewidth]{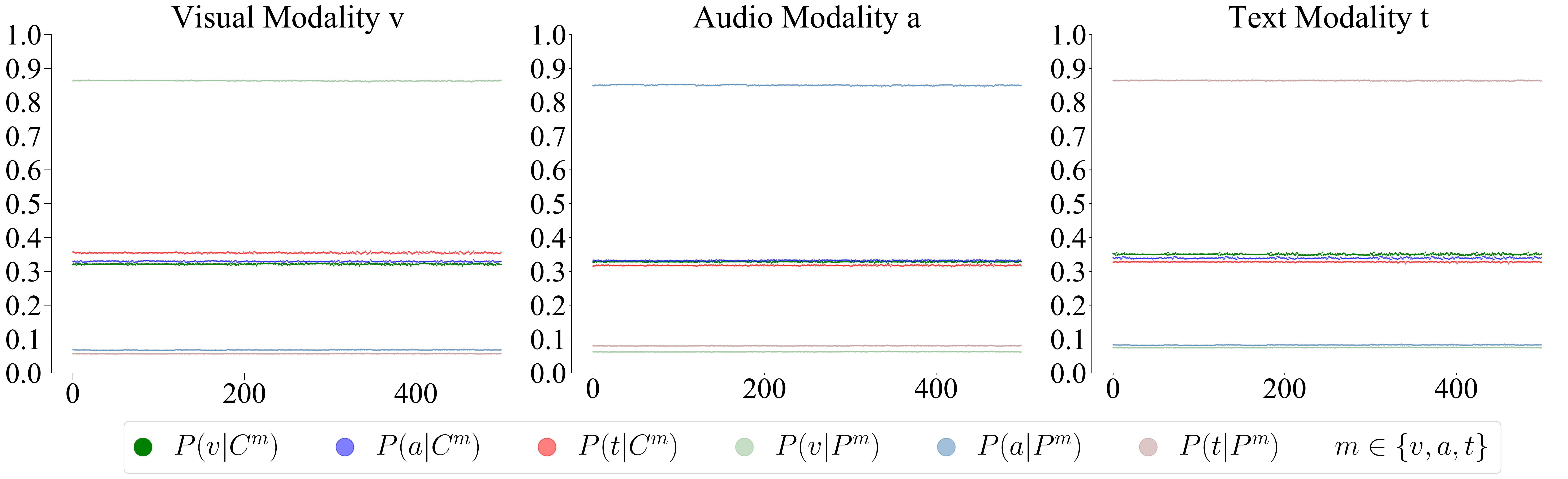} 
\caption{Distribution of private and common representations on different modalities. For modality $m \in \{v, a, t\}$, \\ $[P(v|\bm C^{m}), P(a|\bm C^{m}), P(t|\bm C^{m})] = D(\bm C^{m}; \theta_{D})$,\\ 
$[P(v|\bm P^{m}), P(a|\bm P^{m}), P(t|\bm P^{m})] = D(\bm P^{m}; \theta_{D})$.}
\label{figProb}
\end{figure}
Besides, we display the probability produced by discriminator $D(\cdot; \theta_{D})$ in the AMR.
In Fig.~\ref{figProb}, for each modality m, the probabilities of common representations $P(v|\bm C^{m}), P(a|\bm C^{m}), P(t|\bm C^{m})$ are centered around $0.33$, which is hard to differentiate the source of common modalities.
Contrarily, taking visual modality for example, $P(v|\bm P^{m})$ is higher than $P(a|\bm P^{m})$ and $P(t|\bm P^{m})$ by a large margin, leading to increasingly separable representations.

\noindent\textbf{Visualization of Learned Adversarial Representations}
\\
t-SNE~\cite{van2008visualizing} is adopted to investigate the efficacy of adversarial multi-modal refinement.
We visualize the common and private representations $\bm C^{\{v, a, t\}}$ and $\bm P^{\{v, a, t\}}$ learned without or with adversarial training and orthogonal constraint in aligned CMU-MOSEI.
\begin{figure}[htbp]
\centering
\includegraphics[width=0.9\linewidth]{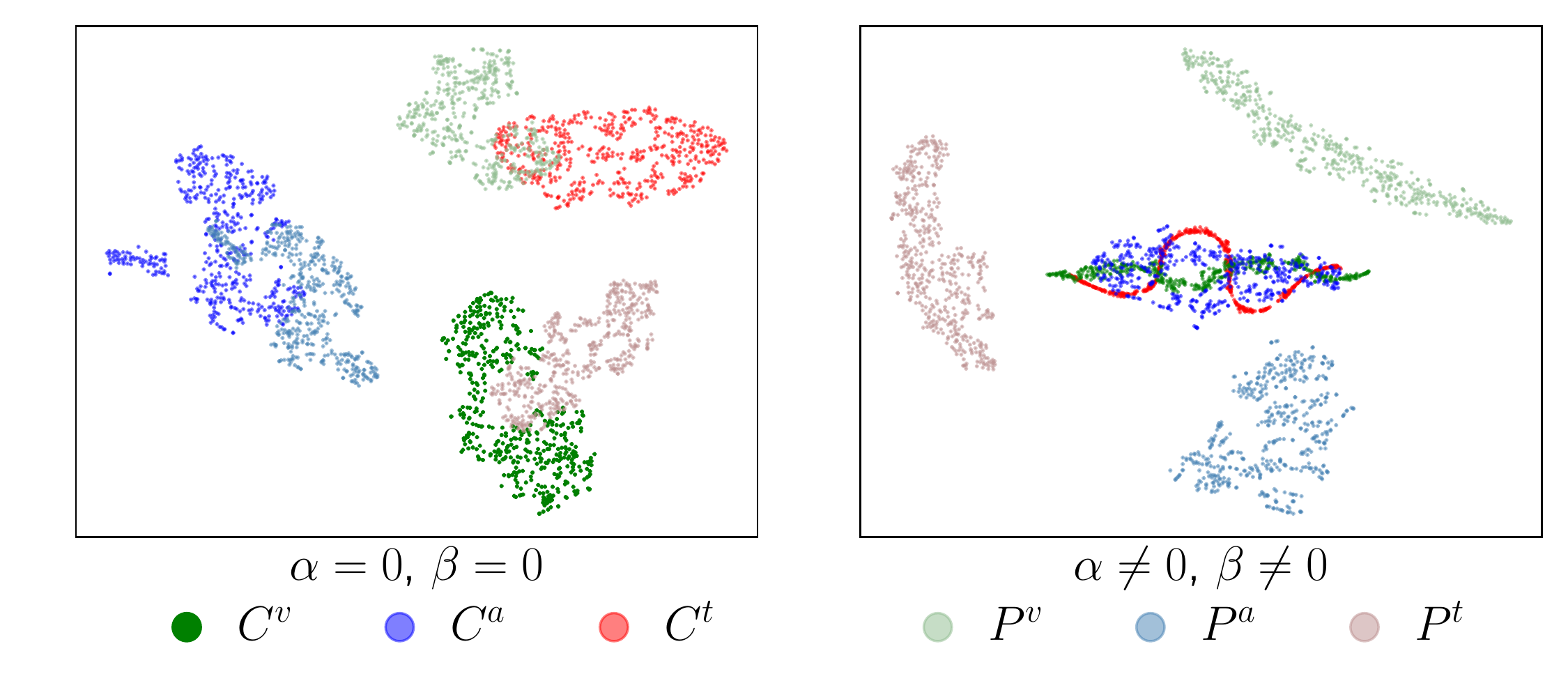} 
\caption{t-SNE visualization of common and private representations in the case without or with adversarial training. The green, blue, red colors represents visual, audio, text modalities respectively.
And dark colors correspond to common parts, while light colors correspond to private parts.}
\label{figtsne}
\end{figure}
As shown in Fig.~\ref{figtsne}, the distributions of $\bm C^{\{v, a, t\}}$ and $\bm P^{\{v, a, t\}}$ are sometimes overlapped in the left subfigure.
Contrarily, in the right subfigure, 1) the distributions of $\bm C^{\{v, a, t\}}$ are mixed together and increasingly blurred, where adversarial training proves effective to align distributions of different modalities and minimize the modality gap;
2) the common latent subspace is separable from each private subspace, where redundant latent representations are penalized with orthogonal constraint.
In all, commonality and specificity of different modalities are well characterized.

\noindent\textbf{Visualization of Learned Label Correlations}
We visualize the learned label correlations $\bm r$ in Eq.~\ref{r} to illustrate the interpretability.
Due to page limit, we only exhibit the results in $4$ attention heads.
\begin{figure}[htbp]
\centering
\includegraphics[width=0.9\linewidth]{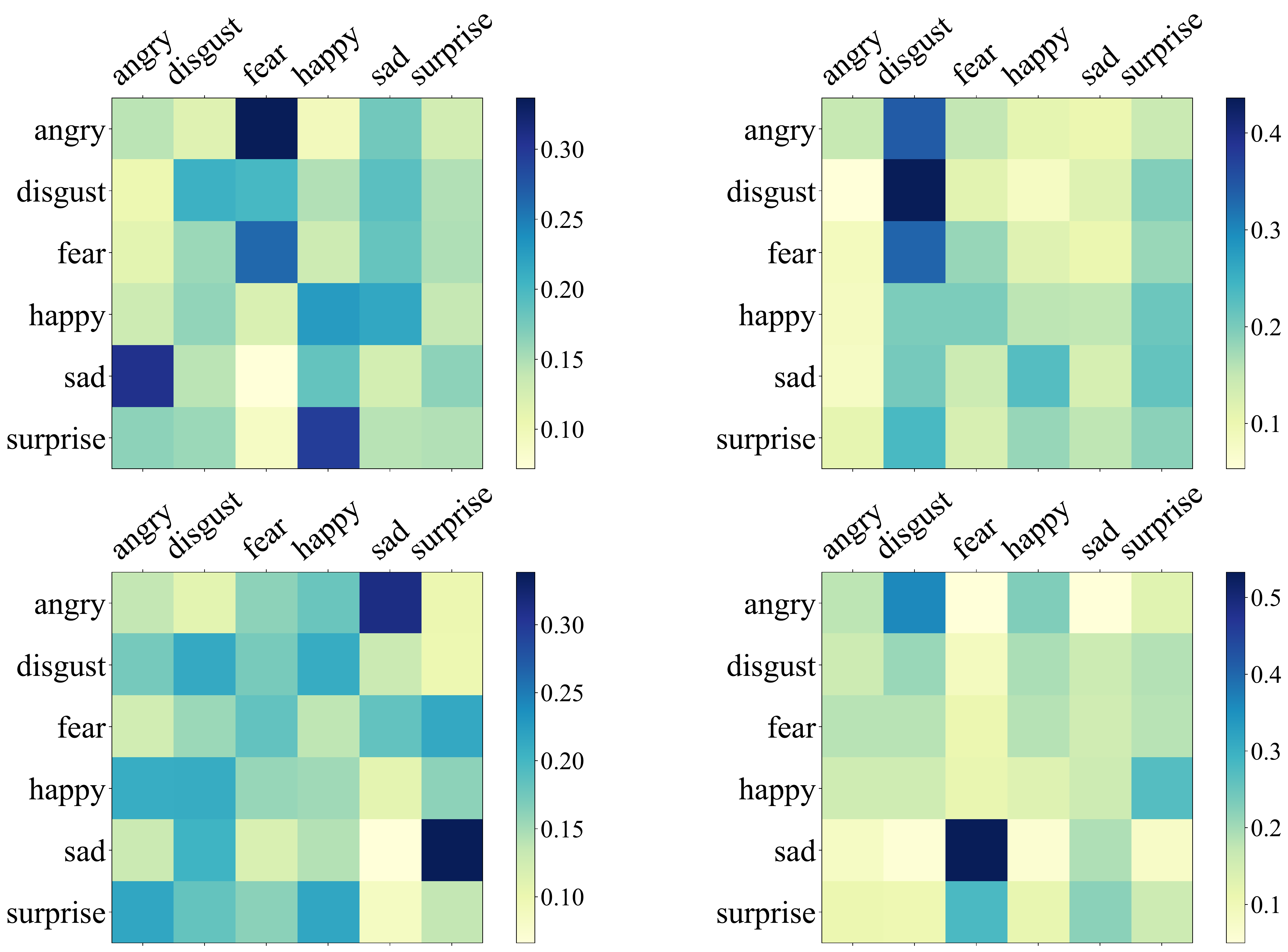} 
\caption{Label correlations visualization, indicating the influence of labels in each row to labels in each column. A higher blue intensity value indicates a stronger correlation.}
\label{figLabelCorr}
\end{figure}
As shown in Figure~\ref{figLabelCorr}, the learned label correlations differ from head to head, which jointly attends to rich semantic information from different perspectives.
From the horizontal view, \textit{angry} is highly correlated with \textit{fear}, \textit{disgust} and \textit{sad} in different heads.
In most cases, \textit{surprise} is highly correlated with \textit{happy}.
All these correlations accord with our intuition.

\section{Conclusion}
In this paper, we propose versaTile multi-modAl learning for multI-labeL emOtion Recognition (TAILOR), consisting of uni-modal extractor, adversarial multi-modal refinement and label-modal alignment.
These modules cooperate closely to refine private and common representations adversarially, fuse multiple modalities in terms of granularity gradually, and leverage label semantics to guide the construction of label-specific representation.
Experimental results and analysis on both aligned and unaligned settings verify effectiveness and generalization of our proposed method. 

\section{Acknowledgments}
This paper is supported by the National Key Research and Development Program of China (Grant No. 2018YFB1403400), the National Natural Science Foundation of China (Grant No. 61876080), the Key Research and Development Program of Jiangsu(Grant No. BE2019105), the Collaborative Innovation Center of Novel Software Technology and Industrialization at Nanjing University.

\bibliography{aaai22}




\end{document}